\documentclass{article}

\usepackage{graphicx}
\usepackage{hyperref}
\usepackage{graphicx}
\usepackage{caption}
\usepackage{subcaption}
\usepackage{color, colortbl}

\begin{document}
\title{Dataset: Rare Event Classification in Multivariate Time Series\thanks{By \href{www.processminer.com}{ProcessMiner, Inc.} }}
\author{Chitta Ranjan\thanks{cranjan@processminer.com} \and
Mahendranath Reddy \and
Markku Mustonen \and
Kamran Paynabar \and
Karim Pourak }
\date{}
% \authorrunning{Ranjan et al.}
% \institute{\email{\{cranjan, markku, kpaynabar, kpourak\}@processminer.com} \\ 715 Peachtree Street N.E. \#100, Atlanta, GA 30308.}
%
\maketitle              % typeset the header of the contribution
\begin{abstract}
A real world dataset is provided from a pulp-and-paper manufacturing industry. The dataset comes from a multivariate time series process. The data contains a rare event of paper break that commonly occurs in the industry. The data contains sensor readings at regular time-intervals (x's) and the event label (y). The primary purpose of the data is thought to be building a classification model for early prediction of rare event. However, it can also be used for multivariate time series data exploration and building other supervised and unsupervised models.

\end{abstract}
\section{Problem}

A multivariate time series (MTS) is produced when multiple correlated streams of data are recorded over time. They are commonly found in manufacturing processes that have several sensors collecting the data in over time. In this problem, we have a similar multivariate time series data from a pulp-and-paper industry with a rare event associated with them. It is an unwanted event in the process —a paper break, in our case — that should be prevented. The objective of the problem is to

\begin{enumerate}
    \item predict the event before it occurs, and
    \item identify the variables that are expected to cause the event (in order to be able to prevent it).
\end{enumerate}

\section{Data}

\begin{figure}
\includegraphics[width=\textwidth]{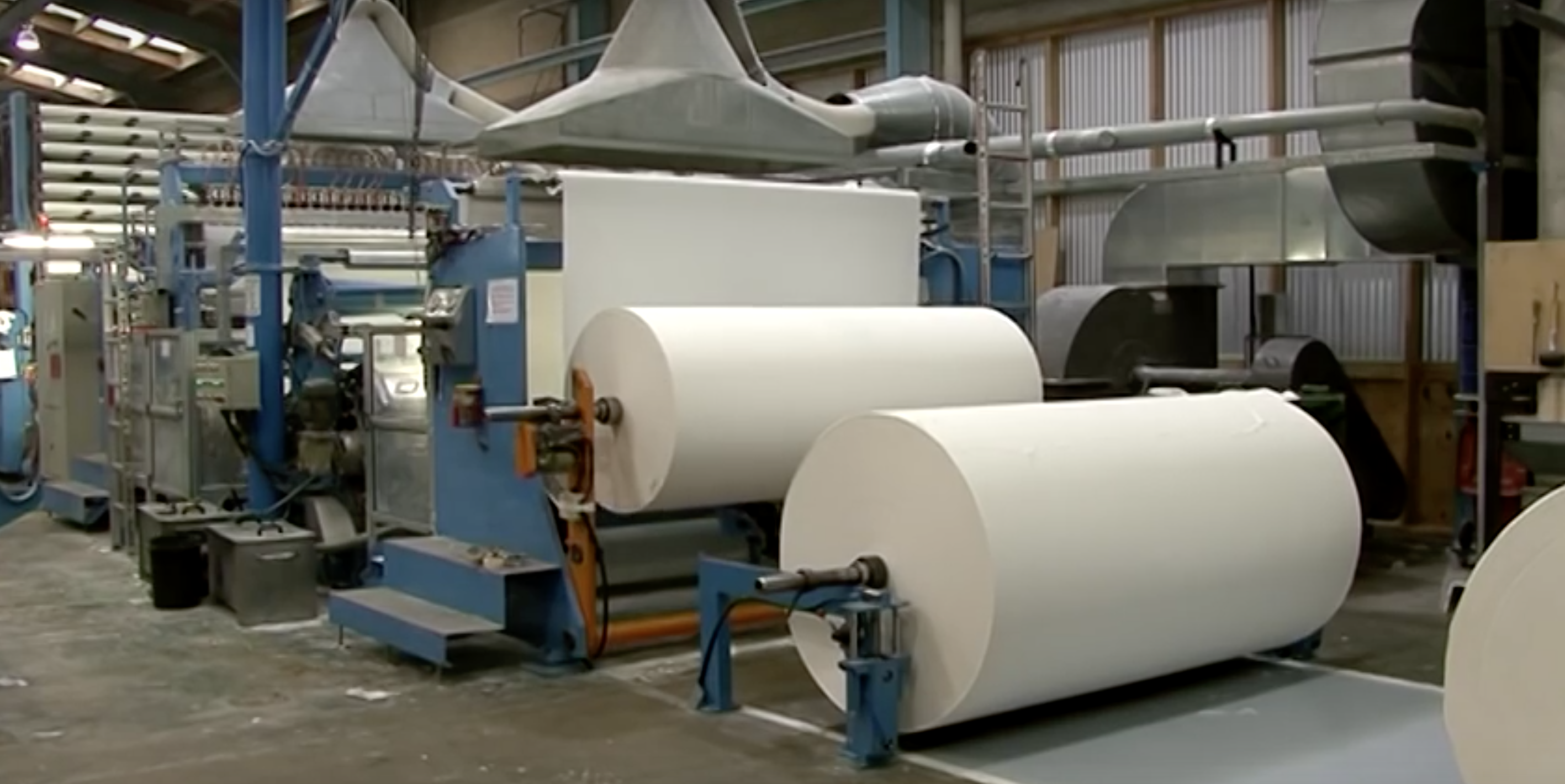}
\caption{A typical paper manufacturing machine.} \label{paper-machine}
\end{figure}

We provide a data from a pulp-and-paper mill. An example of a paper manufacturing machine is shown in Figure~\ref{paper-machine}. These machines are typically several meters long that ingests raw materials at one end and produces reels of paper as shown in the picture.

Several sensors are placed in different parts of the machine along its length and breadth. These sensors measure both raw materials (e.g. amount of pulp fiber, chemicals, etc.) and process variables (e.g. blade type, couch vacuum, rotor speed, etc.).
Paper manufacturing can be viewed as a continuous rolling process. During this process, sometimes the paper breaks. If a break happens, the entire process is stopped, the reel is taken out, any found problem is fixed, and the production is resumed. The resumption can take more than an hour. The cost of this lost production time is significant for a mill. Even a 5\% reduction in the break events will give a significant cost saving for a mill.
The objective of the given problem is to predict such breaks in advance (early prediction) and identify the potential cause(s) to prevent the break. To build such a prediction model, we will use the data collected from the network of sensors in a mill. This is a multivariate time series data with break as the response (a binary variable).

The provided data has,
\begin{itemize}
    \item 18,398 records.
    \item Columns:
    \begin{itemize}
        \item time: the timestamp of the row
        \item y: the binary response variable. There are only 124 rows with y = 1, rest are y = 0.
        \item x1-x61: predictor variables. All the predictors are continuous variables, except x28 and x61. x61 is a binary variable, and x28 is a categorical variable. All the predictors are centered. Besides, the predictors are a mixture of raw materials and process variables. Their descriptions are omitted for data anonymity.
    \end{itemize}
\end{itemize}

Download \href{https://docs.google.com/forms/d/e/1FAIpQLSdyUk3lfDl7I5KYK_pw285LCApc-_RcoC0Tf9cnDnZ_TWzPAw/viewform}{here}.

\section{Challenges}
\textbf{Early classification}

The cause of rare event in the problem is usually instant. For example, a machine failure happens almost instantly after a bolt breaks. Predicting such events before it occurs is thus extremely challenging. However, early detection of a failure is critical to prevent it. 
% Maybe not that relevant
% Similar problem is also found EEG/ECG patient data in hospitals where identifying its abnormalities as soon as possible could offer doctors an emergency alarm and time to save the patient. He et. al. (2015) \cite{he2015} is a good reference to further understand the challenges of an early MTS classification problem.

One straightforward approach you may consider for formulating an early classification problem is: move the class column up by $k$ rows. That is, make $y_{t-k} \leftarrow y_t$, where $k=1,2,\ldots$. In this problem, it is recommended to keep $k$ no more than 2. If $k=1$, we are predicting the event one time unit (equal to 2 mins in the given data) ahead.

\textbf{Feature engineering}

A deep learning approach is usually able to automatically identify important features. However, it requires large amount of data which is a limitation in our problem. Therefore, an appropriate feature engineering becomes critical in developing an effective approach for this problem. Schäfer and Leser (2017) \cite{schafer2017} have elucidated and addressed some major challenges in this regard with a new approach. For example, \textit{multivariate time series} adds large amounts of irrelevant data and noise, and a \textit{high dimensional} problem due to several derived features for each time series in the data.

Schäfer and Leser (2017) have proposed several features. In addition to them, a second derivative of the predictors can also be tried. This is because the first derivative, proposed in Schäfer and Leser (2017), represents gradual change in a variable. A gradual change may not necessarily trigger the event. Sometimes, a sudden change, represented by a second derivative, may cause the event. E.g. in a paper machine there several rollers rotating in sync. A gradual change in the rotational frequency is less likely to cause a break than a sudden out-of-sync change.

Besides, the change in the level of the categorical variable, x28, may be more important than its actual value. This variable is related to the type of paper produced at that time. For this prediction model, it might be more important to capture any change in the paper type instead of the actual type of the paper. May consider adding a feature capturing the change in x28, e.g. $x28_t - x28_{t-1}$.

\textbf{Rare event prediction}

Events, such as failures, are not frequent. Due to this the observed data is usually severely unbalanced. This severely affects the precision and recall of a classification model.

\section{Data Exploration}

A brief exploratory data analysis is done. The distribution of the response variable is provided below.

\begin{figure}
  \centering
  \includegraphics[width=0.5\textwidth]{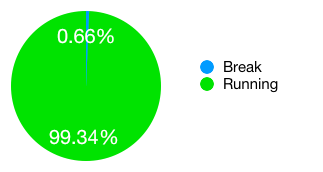}
  \caption{Distribution of the response variable.}
\end{figure}

As shown in the Fig 1, the breaks make less than 1\% of the data. The given data set is \textit{highly imbalanced}. This has to be taken into account during model building as traditional machine learning approaches do not work in such cases.

\begin{figure}
\centering
\begin{subfigure}{.5\textwidth}
  \centering
  \includegraphics[width=1\linewidth]{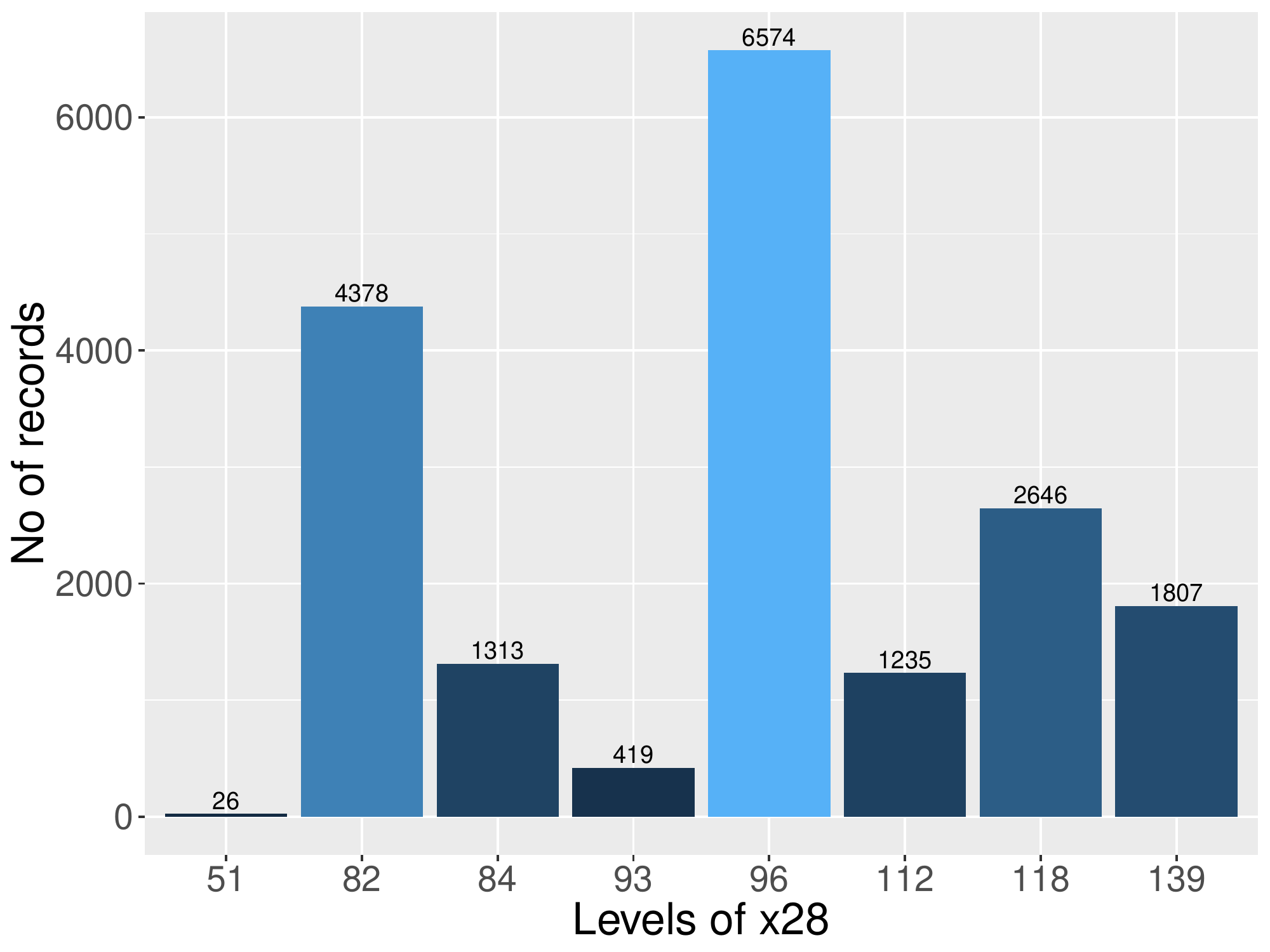}
  \caption{x28}
  \label{fig:sub1}
\end{subfigure}%
\begin{subfigure}{.5\textwidth}
  \centering
  \includegraphics[width=1\linewidth]{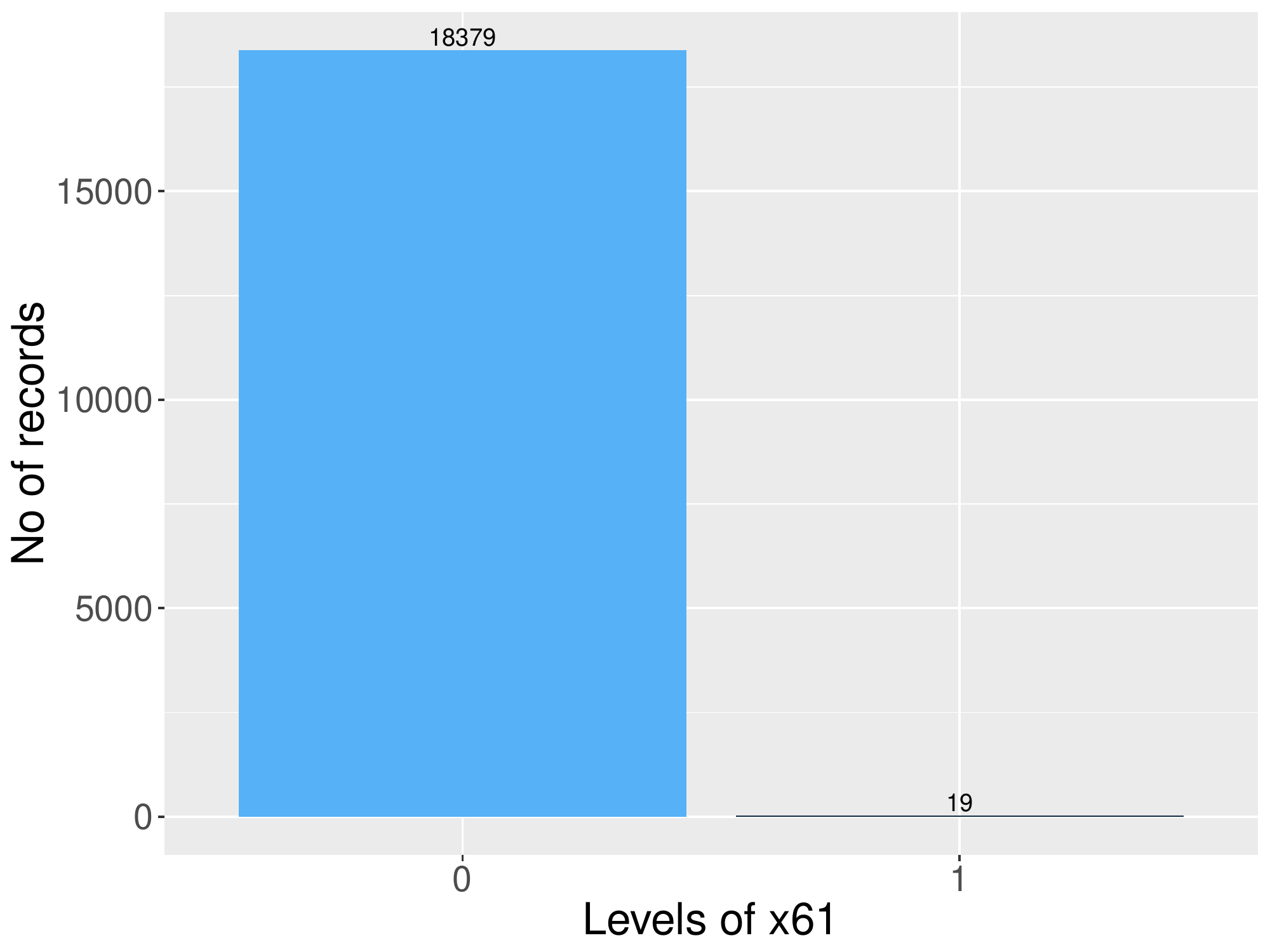}
  \caption{x61}
  \label{fig:sub2}
\end{subfigure}
\caption{Distribution of categorical predictor variables}
\label{fig:test}
\end{figure}

The distribution of categorical predictor variables is shown above. The variable x28 is related to the type of paper produced at that time. For this prediction model, it might be more important to capture any change in the paper type instead of the actual type of the paper.

\section{Modeling approach and results}

The problem given is a rare event early classification problem. The early classification challenge is solved by moving the class column up by $k$ rows. Here, $k$ values of 1 and 2 are tried. If $k=1$, we are predicting the event one time unit (equal to 2 mins in the given data) ahead. The class imbalance is solved using an ensemble built by re sampling the data many times. The idea is to first create new data sets by taking all anomalous data points and adding a subset of normal data points. In addition, feature engineering is done to improve the prediction model.

The dataset is divided into training and test (0.9 training, 0.1 test). XGBoost and AdaBoost models are used to train the prediction model. Initially, models are trained using the given predictor variables. Then, new features are derived and they are used in addition to the given features. Then, first order feature interactions among them in addition to the derived and existing features are used. Finally, frequency domain features based on FFT are used.

\definecolor{Gray}{gray}{0.9}
\definecolor{LightCoral}{rgb}{0.94,0.5,0.5}
\definecolor{PaleGreen}{rgb}{0.6,0.98,0.6}
\definecolor{DeepSkyBlue}{rgb}{0,0.75,1}

Since the given problem is a rare event problem, F1-score is used as the accuracy measure for evaluation. In addition to the F1-score, True Positives (TP), False Positives (FP), True Negatives (TN), False Negatives (FN), Accuracy, Precision, Recall and False Positive Rate (FPR) are reported on the test data.

The best model with the given predictor variables scored a F1-score of 0.081 on the test data. With the addition of derived features, F1-score improved by 40.74\% to 0.114. Precision improved by 51.06\% from 0.047 to 0.071. And, False Positive Rate decreased by 35\% from 0.04 to 0.026.

However, F1-score did not improve with the addition of interaction variables (0.107) or FFT variables (0.099). In both cases, F1-score is better than the baseline model (0.081) but could not beat the model with derived and actual variables (0.114).

\section{Causes of the event}
Our model identified the important variables that lead to a break. This includes derived features of the original variables as well. The derived features result in a significantly powerful model. As shown in Figure 4 below, majority of the important features in our predictive model are derived features.

\begin{figure}
  \centering
  \includegraphics[width=0.5\textwidth]{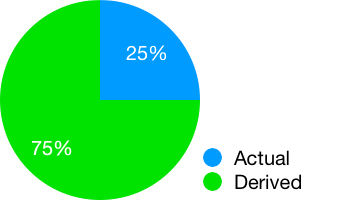}
  \caption{Majority of the important features in the predictive model are some derived features.}
\end{figure}

\begin{figure}
\centering
\begin{subfigure}{.5\textwidth}
  \centering
  \includegraphics[width=1\linewidth]{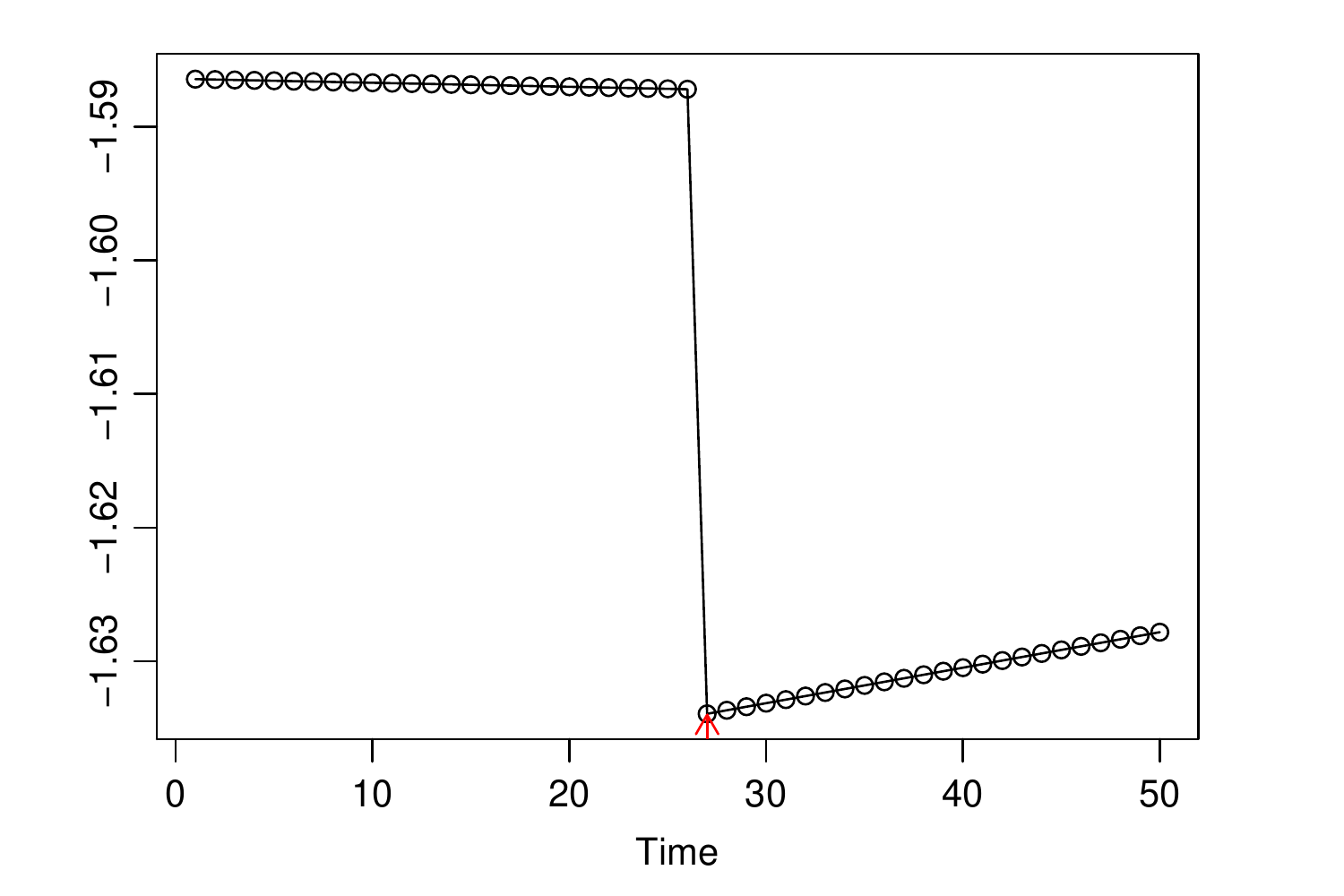}
  \caption{Feature 1}
  \label{fig:sub1}
\end{subfigure}%
\begin{subfigure}{.5\textwidth}
  \centering
  \includegraphics[width=1\linewidth]{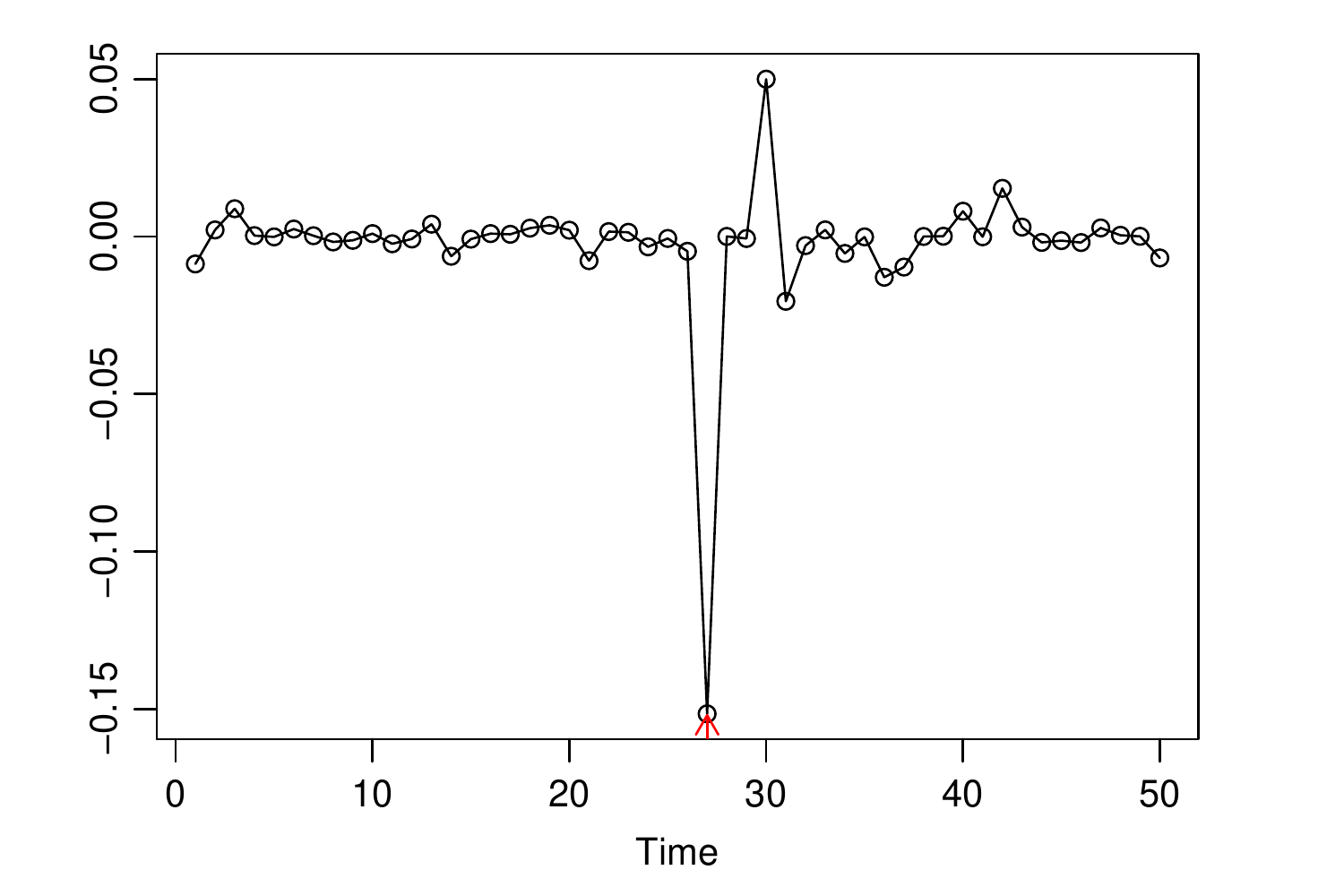}
  \caption{Feature 2}
  \label{fig:sub2}
\end{subfigure}
\caption{Visualization of features in time domain that lead to a break}
\label{fig:test}
\end{figure}

In Figures 5(a) and 5(b) we visualize the shifts in features that lead to a break. We can see a sudden drop in the magnitude of the feature just before the occurrence of a break (red arrow indicates the break event). Note that these are features, which are not necessarily the actual variables.

\begin{figure}
\centering
\begin{subfigure}{.5\textwidth}
  \centering
  \includegraphics[width=1\linewidth]{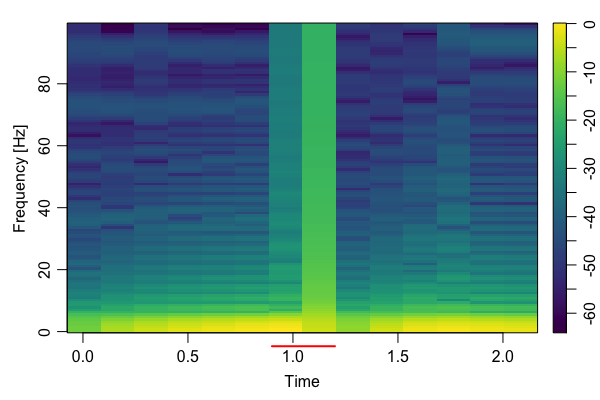}
  \caption{Feature 3}
  \label{fig:sub1}
\end{subfigure}%
\begin{subfigure}{.5\textwidth}
  \centering
  \includegraphics[width=1\linewidth]{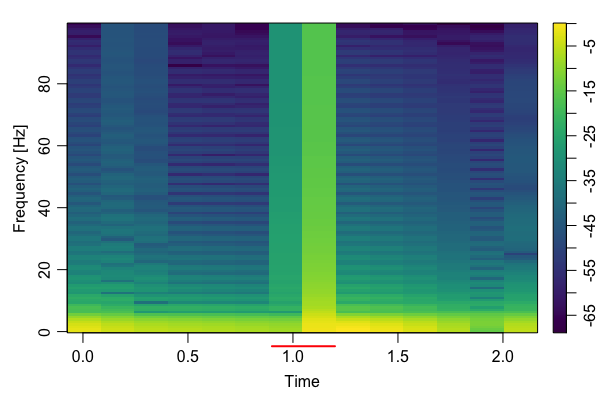}
  \caption{Feature 4}
  \label{fig:sub2}
\end{subfigure}
\caption{Visualization of features in frequency domain that lead to a break}
\label{fig:test}
\end{figure}

In Figures 6(a) and 6(b) we visualize the shifts in few features in frequency domain that lead to a break. The bands underlined in red represents the time just before the occurrence of a break event. We can observe a shift in the amplitude of fft features across nearly all frequencies just before the break event. Note that these are fft features, which are not necessarily the actual variables.

\section{Future directions}
The current best values for F1-score is 0.082 for 2min (one time unit) and 0.114 for 4min (two time units) ahead prediction. This can be further improved by training the prediction model on more amount of data and better feature engineering.

\section{LICENSE}
``Creative Commons Attribution-NonCommercial-ShareAlike 4.0 International Public License'' is included with the data. Under this license, 
\begin{itemize}
    \item users can share the dataset or any publication that uses the data by giving credit to the data provider. The user must cite this paper for the credit.
    \item the dataset cannot be used for any commercial purposes.
    \item users \textbf{can} distribute any additions, transformations or changes to your dataset under this license. However, the same license needs to be added to any redistributed data. Hence, any user of the adapted dataset would likewise need to share their work with this license.
\end{itemize}

\section{Download link}
The data can be downloaded from \href{https://docs.google.com/forms/d/e/1FAIpQLSdyUk3lfDl7I5KYK_pw285LCApc-_RcoC0Tf9cnDnZ_TWzPAw/viewform}{here}. Follow the access instructions on the link.

\end{document}